\documentclass[final]{cvpr}

\usepackage{times}
\usepackage{amsmath, amssymb} 
\usepackage{color}
\usepackage{booktabs}
\usepackage{graphicx}
\usepackage{mathtools}
\usepackage{multirow}
\usepackage{tabularx}

\makeatletter
\@namedef{ver@everyshi.sty}{}
\makeatother
\usepackage{tikz}
\usepackage{pgf, pgfsys, pgffor}
\usepackage{pgfplotstable}
\pgfplotsset{compat=1.15}

\usepackage[pagebackref=true,breaklinks=true,colorlinks,citecolor=blue,linkcolor=red,bookmarks=false]{hyperref}

\def\cvprPaperID{1558} 
\def\confYear{CVPR 2021}

\newcommand{\eat}[1]{}
\newcolumntype{Y}{>{\centering\arraybackslash}X}

\begin{document}
\title{Efficient Regional Memory Network for Video Object Segmentation}

\author{
Haozhe Xie$^{1,2}$\hspace{4 mm}
Hongxun Yao$^1$ \hspace{4 mm}
Shangchen Zhou$^3$ \hspace{4 mm}
Shengping Zhang$^{1,4}$ \hspace{4 mm}
Wenxiu Sun$^{2,5}$\\
$^1$ Harbin Institute of Technology\hspace{4 mm}
$^2$ SenseTime Research and Tetras.AI\\
$^3$ S-Lab, Nanyang Technological University\hspace{4 mm}
$^4$ Peng Cheng Laboratory\hspace{4 mm}
$^5$ Shanghai AI Laboratory\\
\url{https://haozhexie.com/project/rmnet}
}

\maketitle

\begin{abstract}
Recently, several Space-Time Memory based networks have shown that the object cues ({\it e.g.} video frames as well as the segmented object masks) from the past frames are useful for segmenting objects in the current frame.
However, these methods exploit the information from the memory by global-to-global matching between the current and past frames, which lead to mismatching to similar objects and high computational complexity.
To address these problems, we propose a novel local-to-local matching solution for semi-supervised VOS, namely Regional Memory Network (RMNet).
In RMNet, the precise regional memory is constructed by memorizing local regions where the target objects appear in the past frames.
For the current query frame, the query regions are tracked and predicted based on the optical flow estimated from the previous frame.
The proposed local-to-local matching effectively alleviates the ambiguity of similar objects in both memory and query frames, which allows the information to be passed from the regional memory to the query region efficiently and effectively.
Experimental results indicate that the proposed RMNet performs favorably against state-of-the-art methods on the DAVIS and YouTube-VOS datasets.
\end{abstract}

\section{Introduction}

\begin{figure}[!t]
  \resizebox{\linewidth}{!} {
    \includegraphics{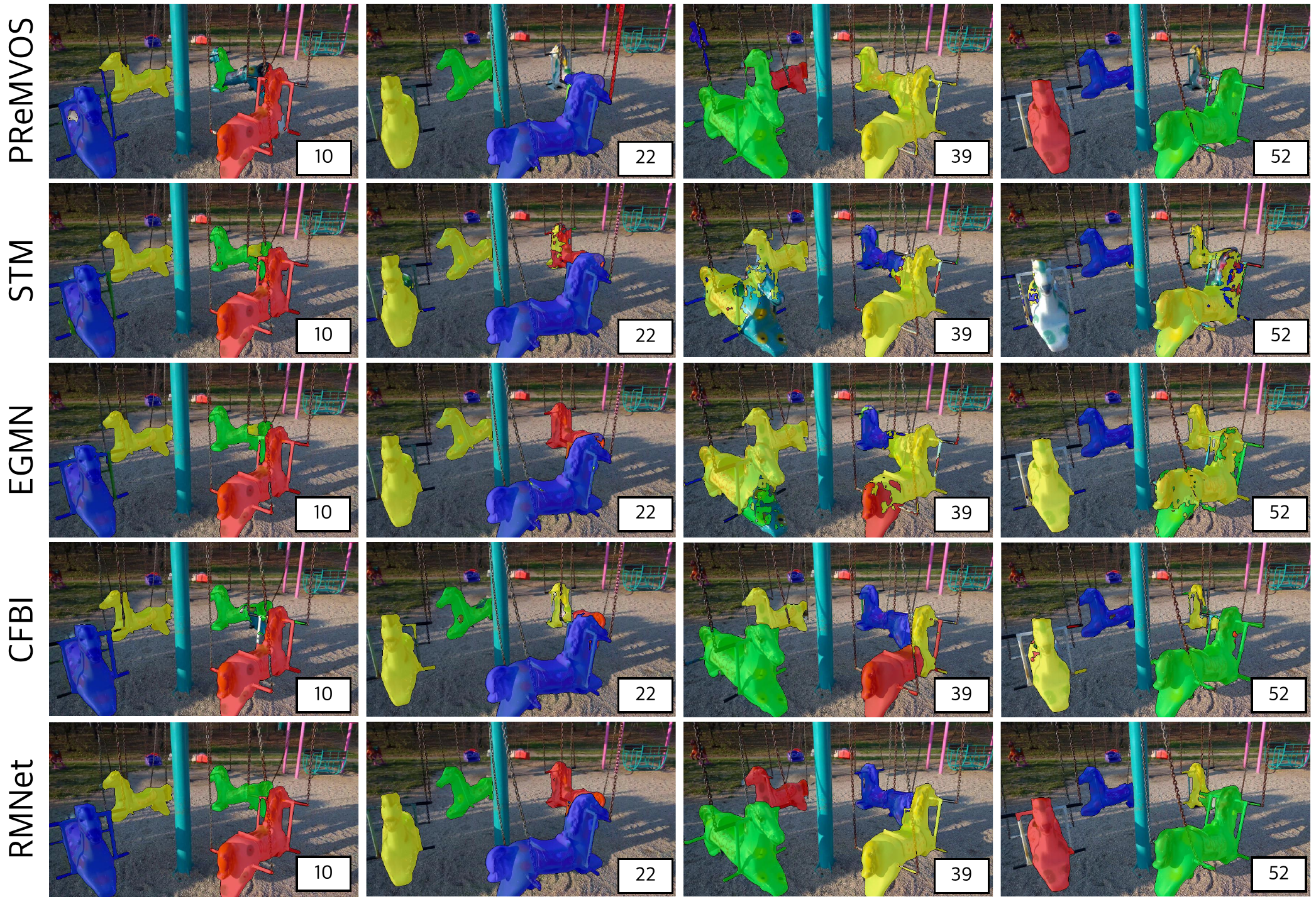}
  }
  \caption{A representative example of video object segmentation on the DAVIS 2017 dataset. Compared to existing methods rely on optical flows ({\it e.g.} PReMVOS~\cite{DBLP:conf/accv/LuitenVL18}) and global feature matching ({\it e.g.} STM~\cite{DBLP:conf/iccv/OhLXK19}, EGMN~\cite{DBLP:conf/eccv/LuWDZSG20}, and CFBI~\cite{DBLP:conf/eccv/YangWY20}), the proposed RMNet is more robust in segmenting similar objects.}
  \label{fig:highlight}
  \vspace{-3 mm}
\end{figure}

Video object segmentation (VOS) is a task of estimating the segmentation masks of class-agnostic objects in a video.
Typically, it can be grouped into two categories: unsupervised VOS and semi-supervised VOS.
The former does not resort to any manual annotation and interaction, while the latter needs the masks of the target objects in the first frame.
In this paper, we focus on the latter.
Even the object masks in the first frame are provided, semi-supervised VOS is still challenging due to object deformation, occlusion, appearance variation, and similar objects confusion.

With the recent advances in deep learning, there has been tremendous progress in semi-supervised VOS.
Early methods~\cite{DBLP:conf/nips/HuHS17,DBLP:conf/accv/LuitenVL18,DBLP:conf/cvpr/PerazziKBSS17} propagate object masks from previous frames using optical flow and then refine the masks with a fully convolutional network.
However, mask propagation usually causes error accumulation, especially when target objects are lost due to occlusions and drifting.
Recently, matching-based methods~\cite{DBLP:conf/cvpr/ChenPMG18,DBLP:conf/eccv/LuWDZSG20,DBLP:conf/iccv/OhLXK19, DBLP:conf/eccv/SeongHK20,DBLP:conf/cvpr/VoigtlaenderCSA19,DBLP:conf/eccv/YangWY20} have attracted increasing attention as a promising solution to semi-supervised VOS.
The basic idea of these methods is to perform global-to-global matching to find the correspondence of target objects between the current and past frames.
Among them, the Space-Time Memory (STM) based approaches~\cite{DBLP:conf/eccv/LuWDZSG20,DBLP:conf/iccv/OhLXK19, DBLP:conf/eccv/SeongHK20} exploit the past frames saved in the memory to better handle object occlusion and drifting.
However, these methods memorize and match features in the regions without target objects, which lead to mismatching to similar objects and high computational complexity.
As shown in Figure~\ref{fig:highlight}, they are less effective to track and distinguish the target objects with similar appearance.

\begin{figure}
  \resizebox{\linewidth}{!} {
    \includegraphics{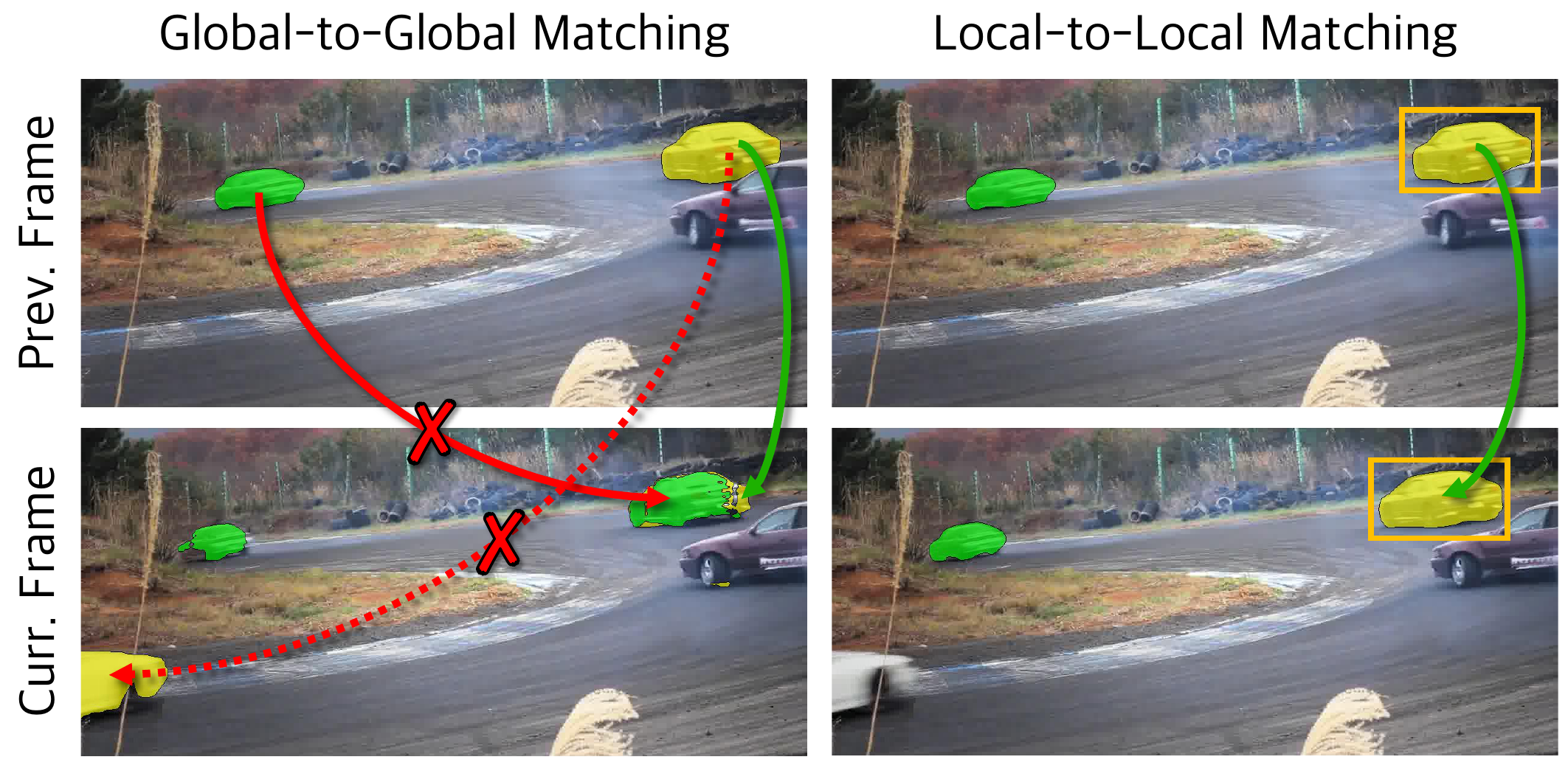}
  }
  \caption{Comparison between global-to-global matching and local-to-local matching. The green and red lines represent correct and incorrect matches, respectively. The proposed local-to-local matching performs feature matching between the local regions containing target objects in past and current frames (highlighted by the yellow bounding box), which alleviates the ambiguity of similar objects.}
  \label{fig:matching-errors}
  \vspace{-2 mm}
\end{figure}

The mismatching in the global-to-global matching can be divided into two categories, as illustrated in Figure~\ref{fig:matching-errors}.
\textbf{(\romannumeral 1)} The target object in the current frame matches to the wrong object in the past frame (solid red line).
\textbf{(\romannumeral 2)} The target object in the past frame matches to the wrong object in the current frame (dotted red line).
The two types of mismatching are caused by unnecessary matching between the regions without target objects in the past and current frames, respectively.
Actually, the target objects appear only in small regions in each frame.
Therefore, it is more reasonable to perform local-to-local matching in the regions containing target objects.

In this paper, we present the Regional Memory Network (RMNet) for semi-supervised VOS.
To better leverage the object cues from the past frames, the proposed RMNet only memorizes the features in the regions containing target objects, which effectively alleviates the mismatching in \textbf{(\romannumeral 1)}.
To track and predict the target object regions for the current frame, we estimate the optical flow from two adjacent frames and then warp the previous object masks to the current frame.
The warped masks provide rough regions for the current frame, which reduces the mismatching in \textbf{(\romannumeral2)}.
Based on the regions in the past and current frames, we present Regional Memory Reader, which performs feature matching between the regions containing target objects.
The proposed Regional Memory Reader is time efficient and effectively alleviates the ambiguity of similar objects.

The main contributions are summarized as follows:
\begin{itemize}
  \vspace{-2.5 mm}
  \item We propose Regional Memory Network (RMNet) for semi-supervised VOS, which memorizes and tracks the regions containing target objects. RMNet effectively alleviates the ambiguity of similar objects.
  \vspace{-2.5 mm}
  \item We present Regional Memory Reader that performs local-to-local matching between object regions in the past and current frames, which reduces the computational complexity.
  \vspace{-2.5 mm}
  \item Experimental results on the DAVIS and YouTube-VOS datasets indicate that the proposed RMNet outperforms the state-of-the-art methods with much faster running speed.
\end{itemize}

\section{Related Work}

\noindent \textbf{Propagation-based Methods.}
Early methods treat video object segmentation as a temporal label propagation problem.
ObjectFlow~\cite{DBLP:conf/cvpr/Tsai0B16}, SegFlow~\cite{DBLP:conf/iccv/ChengTW017} and DVSNet~\cite{DBLP:conf/cvpr/XuFYL18} consider video segmentation and optical flow estimation simultaneously.
To adapt to specific instances, MaskTrack~\cite{DBLP:conf/cvpr/PerazziKBSS17} fine-tunes the network on the first frame during testing.
MaskRNN~\cite{DBLP:conf/nips/HuHS17} predicts the instance-level segmentation of multiple objects from the estimated optical flow and bounding boxes of objects.
CINN~\cite{DBLP:conf/cvpr/BaoW018} introduces a Markov Random Field to  establish spatio-temporal dependencies for pixels.
DyeNet~\cite{DBLP:conf/eccv/LiL18} and PReMVOS~\cite{DBLP:conf/accv/LuitenVL18} combine temporal propagation and re-identification functionalities into a single framework.
Apart from propagating masks with optical flows, object tracking is also widely used in semi-supervised VOS.
FAVOS~\cite{DBLP:conf/cvpr/ChengTHW018} predicts the mask of an object from several tracking boxes of the object parts.
LucidTracker~\cite{DBLP:journals/ijcv/KhorevaBIBS19} synthesizes in-domain data to train a specialized pixel-level video object segmenter.
SAT~\cite{DBLP:conf/cvpr/ChenLYYSQ20} takes advantage of the inter-frame consistency and deals with each target object as a tracklet.
Despite promising results, these methods are not robust to occlusion and drifting, which causes error accumulation during the propagation~\cite{DBLP:conf/cvpr/OhLXK19}.

\noindent \textbf{Matching-based Methods.}
To handle object occlusion and drifting, recent methods perform feature matching to find objects that are similar to the target objects in the rest of the video.
OSVOS~\cite{DBLP:conf/cvpr/CaellesMPLCG17} transfers the generic semantic information to the task of foreground segmentation by fine-tuning the network on the first frame.
RGMP~\cite{DBLP:conf/cvpr/OhLSK18} takes the mask of the previous frame as input, which provides a spatial prior for the current frame.
PML~\cite{DBLP:conf/cvpr/ChenPMG18} learns a pixel-wise embedding using a triplet loss and assigns a label to each pixel by nearest neighbor matching in pixel space to the first frame.
VideoMatch~\cite{DBLP:conf/eccv/HuHS18a} uses a soft matching layer that maps the pixels of the current frame to the first frame in the learned embedding space.
Following PML and VideoMatch, FEELVOS~\cite{DBLP:conf/cvpr/VoigtlaenderCSA19} extends the pixel-level matching mechanism by additionally matching between the current frame and the first frame.
Based on FEELVOS, CFBI~\cite{DBLP:conf/eccv/YangWY20} promotes the results by explicitly considering the feature matching of the target foreground object and the corresponding background. 
STM~\cite{DBLP:conf/iccv/OhLXK19} leverages a memory network to perform pixel-level matching from past frames, which outperforms all previous methods.
Based on STM, KMN~\cite{DBLP:conf/eccv/SeongHK20} applies a Gaussian kernel to reduce the mismatched pixels.
EGMN~\cite{DBLP:conf/eccv/LuWDZSG20} employs an episodic memory network to store frames as nodes and capture cross-frame correlations by edges.
Other STM-based methods~\cite{DBLP:conf/cvpr/XieHXLS20,DBLP:conf/cvpr/ZhangHZP20} use depth estimation and spatial constraint modules to improve the accuracy of STM.
However, matching-based methods perform unnecessary matching in regions without target objects and are less effective to distinguish objects with similar appearance.

\begin{figure*}[!t]
  \resizebox{\linewidth}{!} {
    \includegraphics{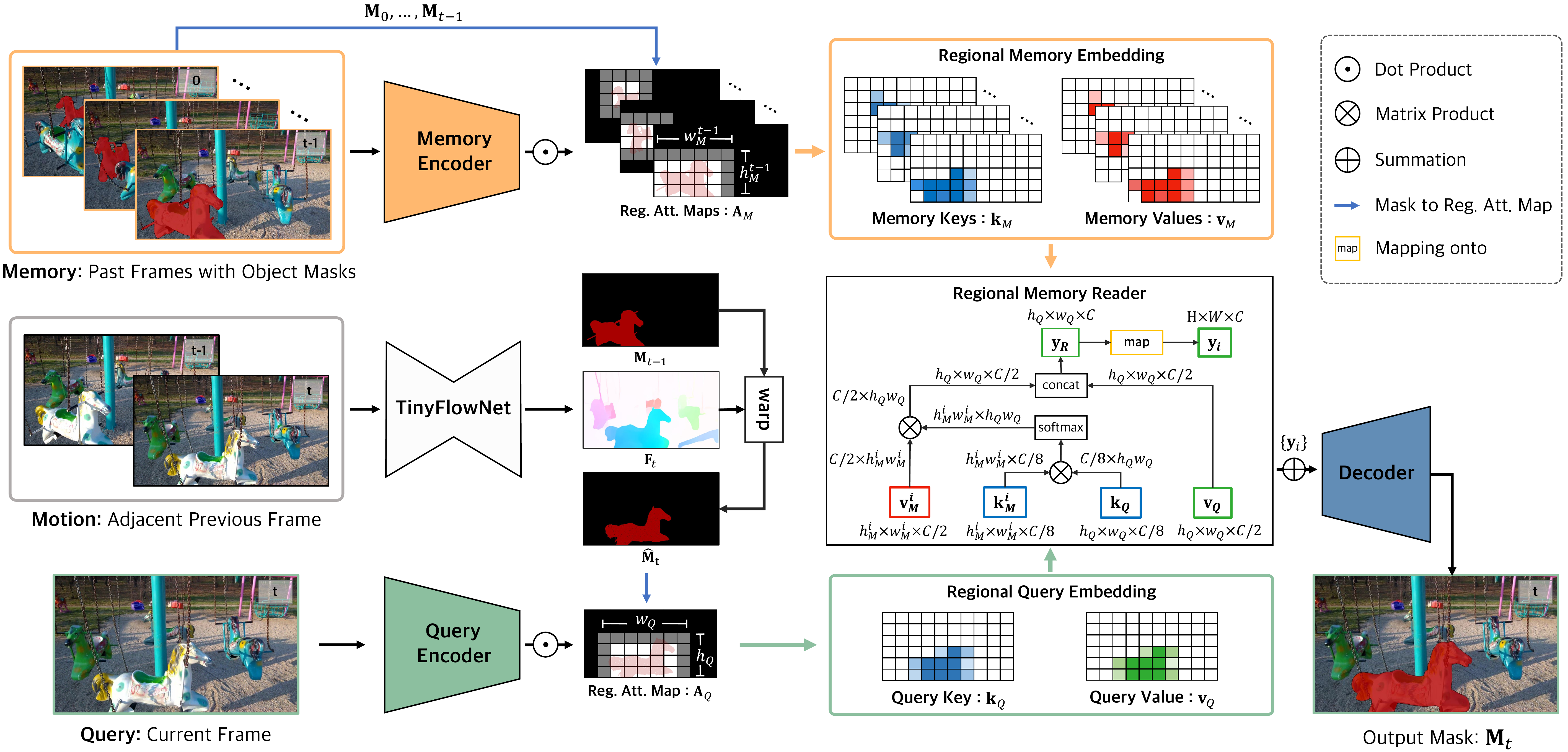}
  }
  \caption{Overview of RMNet.
  The proposed network considers the object motion for the current frame and the object cues from the past frames in memory.
  To alleviate the mismatching to similar objects, the regional memory and query embedding are extracted from the regions containing target objects.
  Regional Memory Reader efficiently performs local-to-local matching only in these regions.
  Note that ``Reg. Att. Map'' denotes ``Regional Attention Map''.}
  \label{fig:overview}
  \vspace{-2 mm}
\end{figure*}

\section{Regional Memory Network}

The architecture of the proposed Regional Memory Network (RMNet) is shown in Figure~\ref{fig:overview}.
As in STM~\cite{DBLP:conf/iccv/OhLXK19}, the current frame is used as the query, and the past frames with the estimated masks are used as the memory.
Different from STM that constructs the global memory and query embedding from all regions, RMNet only embeds the regions containing target objects in the memory and query frames.
The regional memory and query embedding are generated by the dot product of the regional attention maps and feature embedding extracted from the memory and query encoders, respectively.
Both of them consist of a regional \textbf{key} and a regional \textbf{value}.

In STM, the space-time memory reader is employed for global-to-global matching between all pixels in the memory and query frames.
However, Regional Memory Reader in RMNet is proposed for local-to-local matching between the regional memory embedding and query embedding in the regions containing target objects, which alleviates the mismatching to similar objects and also accelerates the computation.
Given the output of Regional Memory Reader, the decoder predicts the object masks for the query frame.

\subsection{Regional Feature Embedding}

\subsubsection{Regional Memory Embedding}
\label{sec:regional-memory-embedding}

Recent Space-Time Memory based methods~\cite{DBLP:conf/eccv/LuWDZSG20,DBLP:conf/iccv/OhLXK19,DBLP:conf/eccv/SeongHK20} construct the global memory embedding for the past frames by using the features of the whole images.
However, the features outside the regions where the target objects appear may lead to the mismatching to the similar objects in the query frame, as shown with the red solid line in Figure~\ref{fig:matching-errors}.
To solve this issue, we present the regional memory embedding that only memorizes the features in the regions containing the target objects.

\noindent \textbf{Mask to Regional Attention Map.}
To generate the regional memory embedding, we apply a regional attention map to the global memory embedding.
At the time step $i$, given the object mask $\mathbf{M}_i \in \mathbb{N}^{H \times W}$ at the feature scale, the regional attention map $\mathbf{A}_i^j \in \mathbb{R}^{H \times W}$ for the $j$-th object is obtained as

\vspace{-4 mm}
\begin{equation}
  \mathbf{A}_i^j(x, y) =
  \begin{cases}
    1, &x_{\min} \le x \le x_{\max}~{\rm and}~y_{\min} \le y \le y_{\max} \\
    0, &{\rm otherwise}
  \end{cases}
\end{equation}
where $(x_{\min}, y_{\min})$ and $(x_{\max}, y_{\max})$ are the top-left and bottom-right coordinates of the bounding box for the target object, which are determined by

\vspace{-4 mm}
\begin{align}
  x_{\min} &= \max((\arg\min_x \mathbf{M}_i(x, y) = j) - \phi, 0)  \nonumber \\
  x_{\max} &= \min((\arg\max_x \mathbf{M}_i(x, y) = j) + \phi, W)  \nonumber \\
  y_{\min} &= \max((\arg\min_y \mathbf{M}_i(x, y) = j) - \phi, 0)  \nonumber \\
  y_{\max} &= \min((\arg\max_y \mathbf{M}_i(x, y) = j) + \phi, H)
\end{align}
where 
$\phi$ denotes the padding of the bounding box, which determines the error tolerance of the estimated masks in the past frames.
Specially, we define $\mathbf{A}_i^j = 0$ if the $j$-th object disappears in $\mathbf{M}_i$.

\noindent \textbf{Regional Memory Key/Value.}
Given the regional attention map $\mathbf{A}_M^j = [\mathbf{A}_0^j, \dots, \mathbf{A}_{t-1}^j]$ of $j$-th object in the memory frames,
the \textbf{key} $\mathbf{k}_M^j$ and \textbf{value} $\mathbf{v}_M^j$ in the regional memory embedding are obtained by the dot product of $\mathbf{A}_M^j$ and the global memory embedding from the memory encoder.

\vspace{-3 mm}
\subsubsection{Regional Query Embedding}

As illustrated by the red dotted line in Figure~\ref{fig:matching-errors}, multiple similar objects in the query frame are easily mismatched in the global-to-global matching.
Similar to the regional memory embedding, we present the regional query embedding that alleviates the mismatching  to the similar objects in the query frame.

\noindent \textbf{Object Mask Tracking.}
To obtain the possible regions of target objects in the current frame, we track and predict a rough mask $\mathbf{\hat{M}}_t^j$.
Specifically, we warp the mask $\mathbf{M}_{t - 1}^j$ of the previous frame with the optical flow $\mathbf{F}_t$ estimated by the proposed TinyFlowNet.

\begin{figure}
  \resizebox{\linewidth}{!} {
    \includegraphics{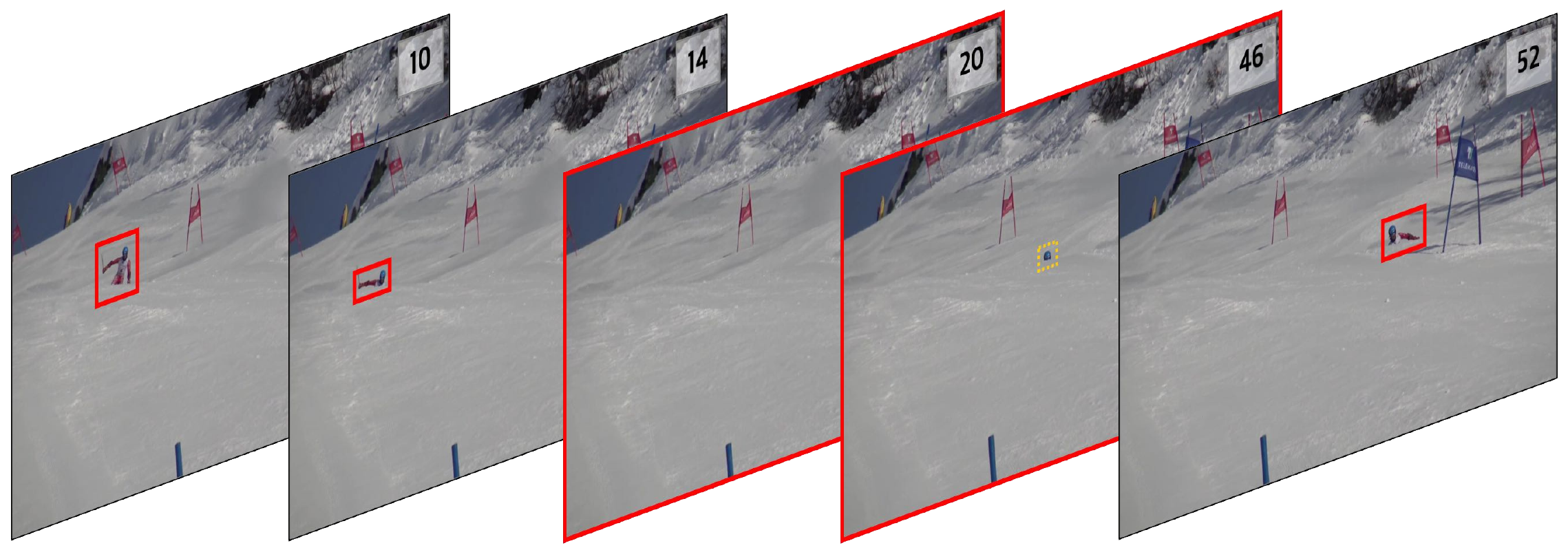}
  }
  \caption{The changes of matching regions for the target object before and after occlusion, which is highlighted by red bounding boxes for each frame.}
  \label{fig:matching-regions}
  \vspace{-2 mm}
\end{figure}

\noindent \textbf{Mask to Regional Attention Map.}
As in regional memory embedding (Sec.~\ref{sec:regional-memory-embedding}), the estimated mask $\mathbf{\hat{M}}_t^j$ is used to generate the regional attention map $\mathbf{A}_Q^j$ for the $j$-th object in the query frame.
To better deal with occlusions, we define $\mathbf{A}_Q^j = 1$ if the number of pixels is lower than a threshold $\eta$, which triggers the global matching for the target object in the query frame.
As illustrated in Figure~\ref{fig:matching-regions}, the matching region is expanded to the whole image when the target object disappears.
Then, it shrinks to the region containing the target object when the object appears again.
This mechanism benefits the optical flow based tracking, which allows the network to perceive the disappearance of objects and makes it more robust to object occlusion.

\noindent \textbf{Regional Query Key/Value.}
Similar to regional memory embedding, the \textbf{key} $\mathbf{k}_Q^j$ and \textbf{value} $\mathbf{v}_Q^j$ in the regional query embedding are obtained by the dot product of $\mathbf{A}_Q^j$ and the global query embedding from the query encoder.

\subsection{Regional Memory Reader}

In STM~\cite{DBLP:conf/iccv/OhLXK19}, the space-time memory reader is proposed to measure the similarities between the pixels in the query frame and memory frames.
Given the embedded \textbf{key} of the memory $\mathbf{k}_M^j = \{k_M^j(\mathbf{p})\} \in \mathbb{R}^{T \cdot H \cdot W \times C / 8}$ and the query $\mathbf{k}_Q^j = \{k_Q^j(\mathbf{q})\} \in \mathbb{R}^{H \cdot W \times C / 8}$ of the $j$-th object, the similarity between $\mathbf{p}$ and $\mathbf{q}$ can be computed as 

\begin{equation}
  s^j(\mathbf{p}, \mathbf{q}) = \exp \left( k_M^j(\mathbf{p}) k_M^j(\mathbf{q})^{\rm T} \right)
\end{equation}
where $C$ and $T$ denote the channels of the embedded key and the number of frames in memory, respectively.
Let $\mathbf{p} = [p_t, p_x, p_y]$ and $\mathbf{q} = [q_x, q_y]$ be the grid cell locations in $\mathbf{k}_M^j$ and $\mathbf{k}_Q^j$, respectively.
Then, the query at position $\mathbf{q}$ retrieves the corresponding \textbf{value} from the memory by

\begin{equation}
  v^j(\mathbf{q}) = \sum_\mathbf{p} \frac{s(\mathbf{p}, \mathbf{q})}{\sum_\mathbf{p} s(\mathbf{p}, \mathbf{q})} v_M^j(\mathbf{p})
\end{equation}
where $\mathbf{v}_M^j = \{v_M^j(\mathbf{p})\} \in \mathbb{R}^{T \times H \times W \times C / 2}$ is the embedded \textbf{value} of the memory.
The output of the space-time memory reader at position $\mathbf{q}$ is

\begin{equation}
  y^j(\mathbf{q}) = \left[ v_Q^j(\mathbf{q}), v^j(\mathbf{q}) \right]
\end{equation}
where $\mathbf{v}_Q^j = \{v_Q^j(\mathbf{q})\} \in \mathbb{R}^{H \times W \times C / 2}$ denotes the embedded \textbf{value} of the query and $[\cdot]$ represents the concatenation.

Based on the regional feature embedding, we propose Regional Memory Reader, which performs local-to-local matching in the regions containing the target objects, as shown in Figure~\ref{fig:overview}.
Compared to the global-to-global memory readers in \cite{DBLP:conf/eccv/LuWDZSG20,DBLP:conf/cvpr/OhLXK19,DBLP:conf/eccv/SeongHK20}, the proposed Regional Memory Reader alleviates the mismatching  to the similar objects in both the memory and query frames.

Let $\mathcal{R}_M^j = \left\{\mathbf{p}\right\}$ and $\mathcal{R}_Q^j = \left\{\mathbf{q}\right\}$ be the feature matching regions of the $j$-th object in the memory and query frames, respectively.
In the global-to-global memory readers, the similarities are derived by a large matrix product.
That is, $\mathcal{R}_M^j$ and $\mathcal{R}_Q^j$ are all locations in the feature embedding of \textbf{key} and \textbf{value}, respectively.
While in the proposed Regional Memory Reader, $\mathcal{R}_M^j$ and $\mathcal{R}_Q^j$ are defined as
\begin{align}
  \mathcal{R}_M^j &= \left\{\mathbf{p} | \mathbf{A}_M^j(\mathbf{p}) \neq 0 \right\} \nonumber
  \\
  \mathcal{R}_Q^j &= \left\{\mathbf{q} | \mathbf{A}_Q^j(\mathbf{q}) \neq 0 \right\}
\end{align}
Therefore, for locations $\mathbf{p} \notin \mathcal{R}_M^j$ or $\mathbf{q} \notin \mathcal{R}_Q^j$, the similarity between $\mathbf{p}$ and $\mathbf{q}$ is defined as

\begin{equation}
  s^j(\mathbf{p}, \mathbf{q}) = 0, 
  \mathbf{p} \notin \mathcal{R}_M^j~{\rm or}~\mathbf{q} \notin \mathcal{R}_Q^j
\end{equation}

Let $h_Q^j$ and $w_Q^j$ be the height and width of the region for the $j$-th object in the query frame, respectively.
$h_M^j$ and $w_M^j$ denote the maximum height and width of the regions in the memory frames, respectively.
Therefore, the time complexity of the space-time memory reader is $\mathcal{O}(T C H^2 W^2)$.
Compared to the space-time memory reader, the proposed Regional Memory Reader is computationally efficient with the time complexity of $\mathcal{O}(T C h_Q^j w_Q^j h_M^j w_M^j)$.
As shown in Figure~\ref{fig:mask-bbox-dist}, $h_Q^j, h_M^j \ll H$ and $w_Q^j, w_M^j \ll W$.
Actually, the space-time memory reader is also a non-local network~\cite{DBLP:conf/cvpr/0004GGH18}, which usually suffers from high computational complexity~\cite{DBLP:conf/iccv/ZhuXBHB19} due to the global-to-global feature matching.
The proposed local-to-local matching enables the time complexity of the memory reader to be significantly reduced.

\subsection{Network Architecture}

\noindent \textbf{TinyFlowNet}.
Compared to existing methods for optical flow estimation~\cite{DBLP:conf/iccv/DosovitskiyFIHH15,DBLP:conf/cvpr/SchusterWUS19,DBLP:conf/cvpr/SunY0K18,DBLP:conf/cvpr/XuRK17}, TinyFlowNet does not use any time-consuming layers such as correlation layer~\cite{DBLP:conf/iccv/DosovitskiyFIHH15}, cost volume layer~\cite{DBLP:conf/cvpr/SunY0K18,DBLP:conf/cvpr/XuRK17}, and dilated convolutional layers~\cite{DBLP:conf/cvpr/SchusterWUS19}.
To reduce the number of parameters of TinyFlowNet, we use small numbers of input and output channels.
Consequently, TinyFlowNet is 1/3 the size of FlowNetS~\cite{DBLP:conf/cvpr/IlgMSKDB17}.
To further accelerate the computation, the input images are downsampled by 2 before fed into TinyFlowNet.

\noindent \textbf{Encoder}.
The memory encoder takes an RGB frame along with the object mask as input, in which the object mask is represented as a single channel probability map between 0 and 1.
The input to the query encoder is only the RGB frame.
Both the memory and query encoders use ResNet50~\cite{DBLP:conf/cvpr/HeZRS16} as the backbone network.
To take a 4-channel tensor, the number of the input channels of the first convolutional layer in the memory encoder is changed to 4.
The first convolutional layer in the query encoder remains unchanged as in ResNet50.
The output \textbf{key} and \textbf{value} features are embedded by two parallel convolutional layers attached to the convolutional layer that outputs a $1/16$ resolution feature with respect to the input image.

\noindent \textbf{Decoder}.
The decoder takes the output of Regional Memory Reader and predicts the object mask for the current frame.
The decoder consists of a residual block and two stacks of refinement modules~\cite{DBLP:conf/iccv/OhLXK19} that gradually upscale the compressed feature map to the size of the input image.

\section{Experiments}

\subsection{Datasets}

\noindent \textbf{DAVIS.}
DAVIS 2016~\cite{DBLP:conf/cvpr/PerazziPMGGS16} is one of the most popular benchmarks for video object segmentation, whose validation set is composed of 20 videos annotated with high-quality masks for individual objects.
DAVIS 2017~\cite{DBLP:preprint/arxiv/1704-00675} is a multi-object extension of DAVIS 2016.
The training and validation sets contain 60 and 30 videos, respectively.

\noindent \textbf{YouTube-VOS.}
YouTube-VOS~\cite{DBLP:conf/eccv/XuYFYYLPCH18} is the latest large-scale dataset for the video object segmentation, which contains 4,453 videos annotated with multiple objects.
Specifically, YouTube-VOS contains 3,471 videos from 65 categories for training and 507 videos with additional 26 unseen categories for validation.

\subsection{Evaluation Metrics}

Following the previous works~\cite{DBLP:conf/iccv/OhLXK19,DBLP:conf/eccv/YangWY20}, we take the region similarity and contour accuracy as evaluation metrics.

\noindent \textbf{Region Similarity $\mathcal{J}$.}
We employ the region similarity $\mathcal{J}$ to measure the region-based segmentation similarity,
which is defined as the intersection-over-union of the estimated segmentation and the ground truth segmentation.
Given an estimated segmentation $M$ and the corresponding ground truth mask $G$, the region similarity $\mathcal{J}$ is defined as

\begin{equation}
  \mathcal{J} = \left|\frac{M \cap G}{M \cup G}\right|
\end{equation}

\noindent \textbf{Contour Accuracy $\mathcal{F}$.}
Let $c(M)$ be the set of the closed contours that delimits the spatial extent of the mask $M$.
The contour points of the estimated mask $M$ and ground truth $G$ are denoted as $c(M)$ and $c(G)$, respectively.
The precision $P_c$ and recall $R_c$ between $c(M)$ and $c(G)$ can be computed by a bipartite graph matching~\cite{DBLP:journals/pami/MartinFM04}.
Therefore, the contour accuracy $\mathcal{F}$ between $c(M)$ and $c(G)$ is defined as

\begin{equation}
  \mathcal{F} = \frac{2 P_c R_c}{P_c + R_c}
\end{equation}

\subsection{Implementation Details}

\begin{table}
  \caption{The quantitative evaluation on the DAVIS 2016 validation set. $^\dag$ indicates using YouTube-VOS for training. The time is measured on an NVIDIA Tesla V100 GPU without I/O time.}
  \vspace{.5 mm}
  \resizebox{\linewidth}{!}{
    \begin{tabularx}{1.08\linewidth}{lcccc}
        \toprule
        Methods       & $\mathcal{J}$ Mean & $\mathcal{F}$ Mean
                      & Avg.               & Time (s) \\
        \midrule
        OnAVOS~\cite{DBLP:conf/bmvc/VoigtlaenderL17}
                      & 0.861      & 0.849      & 0.855      & 0.823 \\
        OSVOS~\cite{DBLP:conf/cvpr/CaellesMPLCG17}
                      & 0.798      & 0.806      & 0.802      & 0.642 \\
        MaskRNN~\cite{DBLP:conf/nips/HuHS17}
                      & 0.807      & 0.809      & 0.808      & - \\
        RGMP~\cite{DBLP:conf/cvpr/OhLSK18}
                      & 0.815      & 0.820      & 0.818      & 0.104 \\
        FAVOS~\cite{DBLP:conf/cvpr/ChengTHW018}
                      & 0.824      & 0.795      & 0.810      & 0.816 \\
        CINN~\cite{DBLP:conf/cvpr/BaoW018}
                      & 0.834      & 0.850      & 0.842      & - \\
        LSE~\cite{DBLP:conf/eccv/CiWW18}
                      & 0.829      & 0.803      & 0.816      & - \\
        VideoMatch~\cite{DBLP:conf/eccv/HuHS18a}
                      & 0.810      & 0.808      & 0.819      & - \\
        PReMVOS~\cite{DBLP:conf/accv/LuitenVL18}
                      & 0.849      & 0.886      & 0.868      & 3.286 \\
        A-GAME $^\dag$~\cite{DBLP:conf/cvpr/JohnanderDBKF19}
                      & 0.822      & 0.820      & 0.821      & 0.258 \\
        FEELVOS $^\dag$~\cite{DBLP:conf/cvpr/VoigtlaenderCSA19}
                      & 0.817      & 0.881      & 0.822      & 0.286 \\
        STM $^\dag$~\cite{DBLP:conf/iccv/OhLXK19}
                      & 0.887      & 0.899      & 0.893      & 0.097 \\
        KMN $^\dag$~\cite{DBLP:conf/eccv/SeongHK20}
                      & \bf{0.895} & \bf{0.915} & \bf{0.905} & - \\
        CFBI $^\dag$~\cite{DBLP:conf/eccv/YangWY20}
                      & 0.883      & 0.905      & 0.894      & 0.156 \\
        \midrule
        RMNet         & 0.806      & 0.823      & 0.815      & \bf{0.084} \\
        RMNet $^\dag$ 
                      & 0.889      & 0.887      & 0.888      & \bf{0.084} \\
        \bottomrule
    \end{tabularx}
  }
  \label{tab:davis2016}
  \vspace{-2 mm}
\end{table}

\begin{table}
  \caption{The quantitative evaluation on the DAVIS 2017 validation set. $^\dag$ indicates using YouTube-VOS for training.}
  \vspace{.5 mm}
  \begin{tabularx}{\linewidth}{lYYY}
    \toprule
    Methods       & $\mathcal{J}$ Mean & $\mathcal{F}$ Mean
                  & Avg. \\
    \midrule
    OnAVOS~\cite{DBLP:conf/bmvc/VoigtlaenderL17}
                  & 0.645      & 0.713      & 0.679 \\
    OSMN~\cite{DBLP:conf/cvpr/YangWXYK18}
                  & 0.525      & 0.571      & 0.548 \\
    OSVOS~\cite{DBLP:conf/cvpr/CaellesMPLCG17}
                  & 0.566      & 0.618      & 0.592 \\
    RGMP~\cite{DBLP:conf/cvpr/OhLSK18}
                  & 0.648      & 0.686      & 0.632 \\
    FAVOS~\cite{DBLP:conf/cvpr/ChengTHW018}
                  & 0.546      & 0.618      & 0.582 \\
    CINN~\cite{DBLP:conf/cvpr/BaoW018}
                  & 0.672      & 0.742      & 0.707 \\
    VideoMatch~\cite{DBLP:conf/eccv/HuHS18a}
                  & 0.565      & 0.682      & 0.624 \\
    PReMVOS~\cite{DBLP:conf/accv/LuitenVL18}
                  & 0.739      & 0.817      & 0.778 \\
    A-GAME $^\dag$~\cite{DBLP:conf/cvpr/JohnanderDBKF19}
                  & 0.672      & 0.727      & 0.700 \\
    FEELVOS $^\dag$~\cite{DBLP:conf/cvpr/VoigtlaenderCSA19}
                  & 0.691      & 0.740      & 0.716 \\
    STM $^\dag$~\cite{DBLP:conf/iccv/OhLXK19}
                  & 0.792      & 0.843      & 0.818 \\
    KMN $^\dag$~\cite{DBLP:conf/eccv/SeongHK20}
                  & 0.800      & 0.856      & 0.828 \\
    EGMN $^\dag$~\cite{DBLP:conf/eccv/LuWDZSG20} 
                  & 0.800      & 0.859      & 0.829 \\
    CFBI $^\dag$~\cite{DBLP:conf/eccv/YangWY20}
                  & 0.791      & 0.846      & 0.819 \\
    \midrule
    RMNet         & 0.728      & 0.772      & 0.750 \\
    RMNet $^\dag$ 
                  & \bf{0.810} & \bf{0.860} & \bf{0.835} \\
    \bottomrule
  \end{tabularx}
  \label{tab:davis2017-val}
  \vspace{-2 mm}
\end{table}

\begin{table}
  \caption{The quantitative evaluation on the DAVIS 2017 test-dev set.}
  \vspace{.5 mm}
  \begin{tabularx}{\linewidth}{lYYY}
    \toprule
    Methods       & $\mathcal{J}$ Mean & $\mathcal{F}$ Mean
                  & Avg. \\
    \midrule
    OnAVOS~\cite{DBLP:conf/bmvc/VoigtlaenderL17}
                  & 0.534      & 0.596      & 0.565 \\
    OSMN~\cite{DBLP:conf/cvpr/YangWXYK18}
                  & 0.377      & 0.449      & 0.413 \\
    RGMP~\cite{DBLP:conf/cvpr/OhLSK18}
                  & 0.513      & 0.544      & 0.529 \\
    PReMVOS~\cite{DBLP:conf/accv/LuitenVL18}
                  & 0.675      & 0.757      & 0.716 \\
    FEELVOS~\cite{DBLP:conf/cvpr/VoigtlaenderCSA19}
                  & 0.552      & 0.605      & 0.578 \\
    STM~\cite{DBLP:conf/iccv/OhLXK19}
                  & 0.680      & 0.740      & 0.710 \\
    CFBI~\cite{DBLP:conf/eccv/YangWY20}
                  & 0.711      & \bf{0.785} & 0.748 \\
    \midrule
    RMNet         & \bf{0.719} & 0.781      & \bf{0.750} \\
    \bottomrule
  \end{tabularx}
  \label{tab:davis2017-test-dev}
  \vspace{-2 mm}
\end{table}

We implement the network using PyTorch~\cite{DBLP:conf/nips/AdamSSGEZZALA19} and CUDA.
All models are optimized using the Adam optimizer~\cite{DBLP:conf/iclr/KingmaB15} with $\beta_1$ = 0.9 and $\beta_2$ = 0.999.
Following~\cite{DBLP:conf/iccv/OhLXK19,DBLP:conf/eccv/SeongHK20}, the network is trained in two phases.
First, it is pretrained on the synthetic dataset generated by applying random affine transforms to a static image with different parameters.
Then, it is fine-tuned on DAVIS and YouTube-VOS. 
%
The parameters $\phi$ and $\eta$ are set to 4 and 10, respectively. 
For all experiments, the network is trained with a batch size of 4 on two NVIDIA Tesla V100 GPUs.
All batch normalization layers are fixed during training and testing.
The initial learning rate is set to $10^{-5}$ and the optimization is set to stop after 200 epochs.

\subsection{Video Object Segmentation on DAVIS}

\noindent \textbf{Single object.}
We compare proposed RMNet  with state-of-the-art methods on the validation set of the DAVIS 2016 dataset for single-object video segmentation.
DAVIS contains only a small number of videos, which leads to overfitting and affects the generalization ability.
Following the latest works~\cite{DBLP:conf/cvpr/ChengTHW018,DBLP:conf/iccv/OhLXK19,DBLP:conf/eccv/SeongHK20,DBLP:conf/eccv/YangWY20}, we also present the results trained with additional data from YouTube-VOS, which are denoted as $^\dag$ in Table~\ref{tab:davis2016}.
The experimental results indicate that the proposed RMNet is comparable to other competitive methods but with a faster inference speed.

\noindent \textbf{Multiple objects.}
To evaluate the performance of multi-object video segmentation, we test the proposed RMNet on the DAVIS 2017 benchmark.
We report the performance of the val set of DAVIS 2017 in Table~\ref{tab:davis2017-val}, which shows that RMNet outperforms all competitive methods.
With additional YouTube-VOS data, RMNet archives a better accuracy and outperforms all state-of-the-art methods.
We also evaluate RMNet on the test-dev set of DAVIS 2017, which is much more challenging than the val set.
As shown in Table~\ref{tab:davis2017-test-dev}, RMNet surpasses the state-of-the-art methods.
The qualitative results on DAVIS 2017 are shown in Figure~\ref{fig:highlight}.

\subsection{Video Object Segmentation on YouTube-VOS}

\begin{table}
  \caption{The quantitative evaluation on the YouTube-VOS validation set (2018 version). The results of other methods are directly copied from~\cite{DBLP:conf/iccv/OhLXK19,DBLP:conf/eccv/YangWY20}.}
  \vspace{.5 mm}
  \begin{tabularx}{\linewidth}{lYYYYY}
    \toprule
    \multirow{2}{*}{Methods} &
    \multicolumn{2}{c}{Seen} & \multicolumn{2}{c}{Unseen} &
    \multirow{2}{*}{Avg.} \\
    \noalign{\smallskip} \cline{2-5} \noalign{\smallskip}
                  & $\mathcal{J}$ & $\mathcal{F}$
                  & $\mathcal{J}$ & $\mathcal{F}$ \\
    \midrule
    OnAVOS~\cite{DBLP:conf/bmvc/VoigtlaenderL17}
                  & 0.601      & 0.627      & 0.466
                  & 0.514      & 0.552 \\
    OSMN~\cite{DBLP:conf/cvpr/YangWXYK18}
                  & 0.600      & 0.601      & 0.406
                  & 0.440      & 0.512 \\
    OSVOS~\cite{DBLP:conf/cvpr/CaellesMPLCG17}
                  & 0.598      & 0.605      & 0.542
                  & 0.607      & 0.588 \\
    RGMP~\cite{DBLP:conf/cvpr/OhLSK18}
                  & 0.595      & -          & 0.452
                  & -          & 0.538 \\
    BoLTVOS~\cite{DBLP:preprint/arxiv/1904-04552}
                  & 0.716      & -          & 0.643
                  & -          & 0.711 \\
    PReMVOS~\cite{DBLP:conf/accv/LuitenVL18}
                  & 0.714      & 0.759      & 0.565
                  & 0.637      & 0.669 \\
    A-GAME~\cite{DBLP:conf/cvpr/JohnanderDBKF19}
                  & 0.678      & -          & 0.608
                  & -          & 0.661 \\
    STM~\cite{DBLP:conf/iccv/OhLXK19}
                  & 0.797      & 0.842      & 0.728
                  & 0.809      & 0.794 \\
    KMN~\cite{DBLP:conf/eccv/SeongHK20}
                  & 0.814      & 0.856      & 0.753
                  & 0.833      & 0.814 \\
    EGMN~\cite{DBLP:conf/eccv/LuWDZSG20} 
                  & 0.807      & 0.851      & 0.740
                  & 0.809      & 0.802 \\
    CFBI~\cite{DBLP:conf/eccv/YangWY20}
                  & 0.811      & \bf{0.858} & 0.753 
                  & \bf{0.834} & 0.814 \\
    \midrule
    RMNet         & \bf{0.821} & 0.857      & \bf{0.757}
                  & 0.824      & \bf{0.815} \\
    \bottomrule
  \end{tabularx}
  \label{tab:youtubevos}
  \vspace{-2 mm}
\end{table}

\begin{figure*}
  \resizebox{\linewidth}{!} {
    \includegraphics{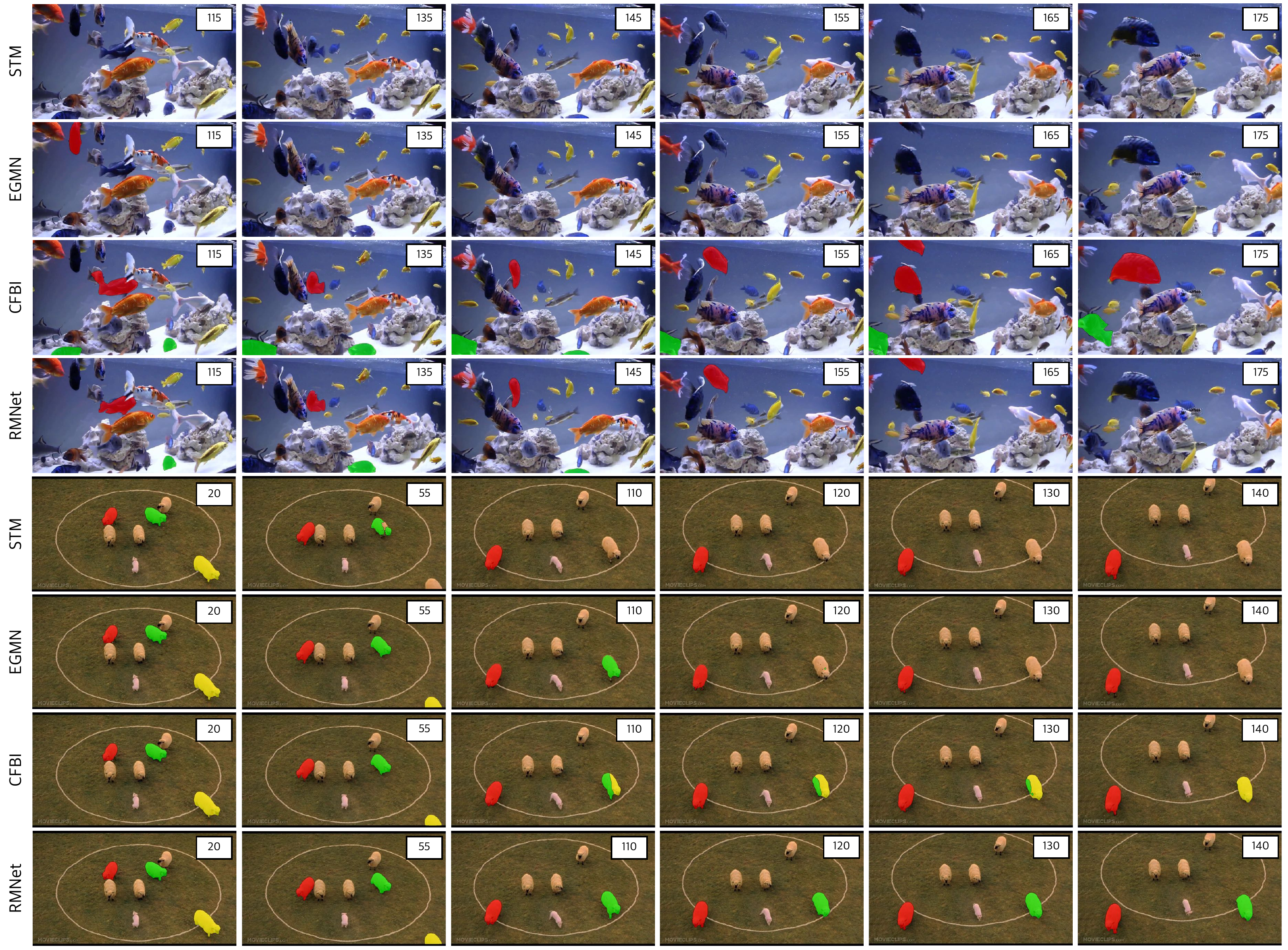}
  }
  \caption{The challenging examples of multi-object video segmentation on the YouTube-VOS validation set (2018 version). All methods are tested at 720p without test-time augmentation. For the first video, both STM~\cite{DBLP:conf/iccv/OhLXK19} and EGMN~\cite{DBLP:conf/eccv/LuWDZSG20} fail to segment the target objects before the 115-th frame.}
  \label{fig:youtubevos}
  \vspace{-2 mm}
\end{figure*}

Following the latest works~\cite{DBLP:conf/iccv/OhLXK19,DBLP:conf/eccv/SeongHK20}, we compare the proposed RMNet to the state-of-the-art methods on the YouTube-VOS validation set (2018 version).
As shown in Table~\ref{tab:youtubevos}, RMNet achieves an average score of 0.815, which outperforms other methods.
The qualitative results on the YouTube-VOS dataset are shown in Figure~\ref{fig:youtubevos}, which demonstrate that RMNet is more effective in distinguishing similar objects and performs better in segmenting small objects.

\subsection{Ablation Study}

To demonstrate the effectiveness of each component in the proposed RMNet, we conduct the ablation study on the DAVIS 2017 val set.

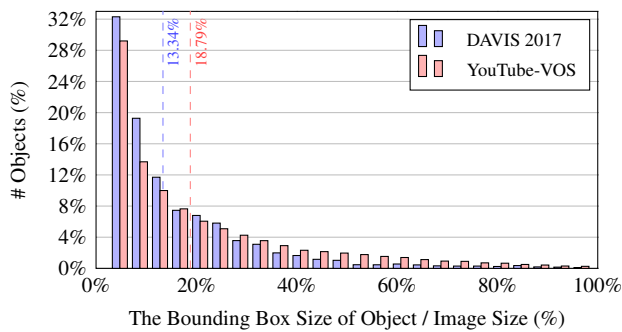
\begin{figure}
  \begin{tikzpicture}
    \begin{axis}[
      ybar,
      width = .99\linewidth,
      height = .6\linewidth,
      bar width = 1 mm,
      bar shift = 0,
      xmin = 0,
      xmax = 1,
      ymin = 0,
      ymax = 0.33,
      legend style={
        at = {(0.805, 0.97)},
        anchor = north,
        column sep = 0.25cm,
        font = \fontsize{7}{7}\selectfont
      },
      legend cell align = left,
      xlabel = {The Bounding Box Size of Object / Image Size (\%)},
      xlabel style = {
        font = \fontsize{8}{8}\selectfont
      },
      ylabel = {\# Objects (\%)},
      ylabel style = {
        font = \fontsize{8}{8}\selectfont
      },
      tickwidth = 0pt,
      xtick = {0, 0.2, 0.4, 0.6, 0.8, 1.0},
      xticklabels = {0\%, 20\%, 40\%, 60\%, 80\%, 100\%},
      xticklabel style={
        font = \fontsize{8}{8}\selectfont,
        align = center
      },
      ytick = {0, 0.04, 0.08, 0.12, 0.16, 0.20, 0.24, 0.28, 0.32},
      yticklabels = {0\%, 4\%, 8\%, 12\%, 16\%, 20\%, 24\%, 28\%, 32\%},
      yticklabel style={
        font = \fontsize{8}{8}\selectfont,
        align = center
      },
      ymajorgrids = true,
    ]
    
    \draw[blue!50, dashed] (0.13339, 0) -- (0.13339, 0.36);
    
    \node[blue!75, rotate=90] at (0.154, 0.29) {\fontsize{6}{6}\selectfont 13.34\%};
    
    \draw[red!50, dashed] (0.18789, 0) -- (0.18789, 0.36);
    
    \node[red!75, rotate=90] at (0.21, 0.29) {\fontsize{6}{6}\selectfont 18.79\%};

    \addplot[fill=blue!30]
    coordinates {
      (0.04, 0.32304822) (0.08, 0.19273823) (0.12, 0.11703502) (0.16, 0.07455511)
      (0.20, 0.0679535 ) (0.24, 0.05812285) (0.28, 0.03566303) (0.32, 0.0309271 )
      (0.36, 0.01987658) (0.40, 0.01650402) (0.44, 0.01162457) (0.48, 0.01033295)
      (0.52, 0.00466418) (0.56, 0.00459242) (0.60, 0.00559701) (0.64, 0.00444891)
      (0.68, 0.00322905) (0.72, 0.00294202) (0.76, 0.00294202) (0.80, 0.00243972)
      (0.84, 0.00358783) (0.88, 0.00186567) (0.92, 0.00157865) (0.96, 0.00093284) 
    };
    \addlegendentry{DAVIS 2017};

    \addplot[fill=red!30, bar shift= 1mm]
    coordinates {
      (0.04 , 0.292026  ) (0.08 , 0.13684111) (0.12 , 0.09994386) (0.16 , 0.07642567)
      (0.20 , 0.06055748) (0.24 , 0.0508783 ) (0.28 , 0.04256542) (0.32 , 0.03559181)
      (0.36 , 0.02918637) (0.40 , 0.02323411) (0.44 , 0.02148901) (0.48 , 0.01964246)
      (0.52 , 0.01770121) (0.56 , 0.0153406 ) (0.60 , 0.01394723) (0.64 , 0.01127547)
      (0.68 , 0.00934099) (0.72 , 0.00901632) (0.76 , 0.00706831) (0.80 , 0.00662865)
      (0.84 , 0.00503237) (0.88 , 0.00426804) (0.92 , 0.00300319) (0.96 , 0.00261088) 
    };
    \addlegendentry{YouTube-VOS};
    \end{axis}
  \end{tikzpicture}
  \caption{The long tail distribution of the area ratio of the bounding boxes of target objects on the training set of the DAVIS 2017 and YouTube-VOS datasets. The vertical dashed lines denote the mean value of the area ratio of bounding boxes for both datasets.}
  \label{fig:mask-bbox-dist}
  \vspace{-4 mm}
\end{figure}

\begin{figure*}
  \resizebox{\linewidth}{!} {
    \includegraphics{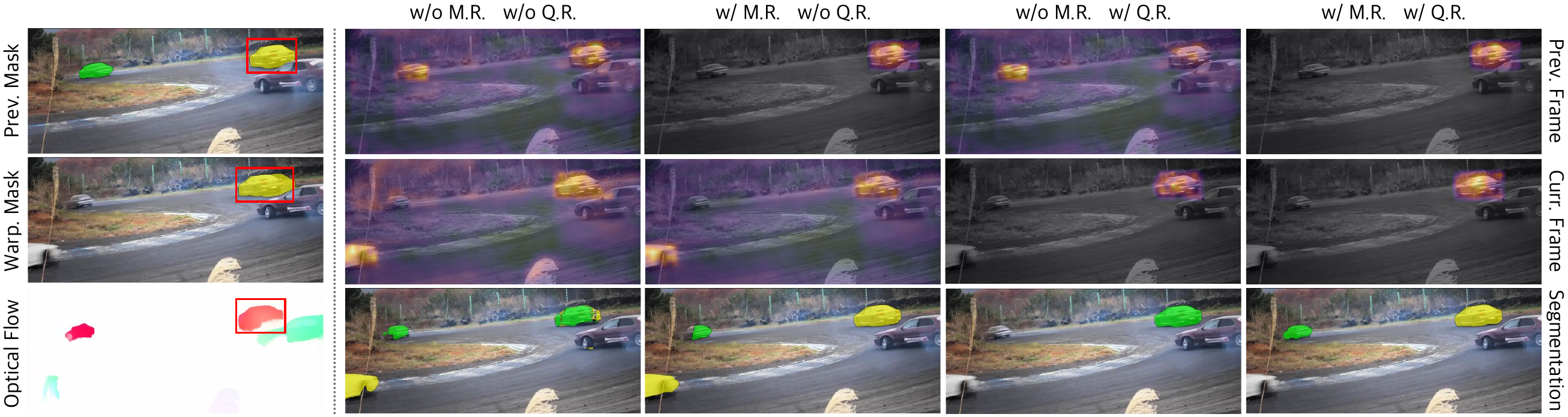}
  }
  \caption{The similarity scores of the target object in the previous and current frames. The target object is highlighted by a red bounding box. ``M.R.'' and ``Q.R.'' denote for ``Memory Region'' and ``Query Region'' in the regional memory and query embedding, respectively.}
  \label{fig:ablation-regional-memory-reader}
  \vspace{-4 mm}
\end{figure*}

\noindent \textbf{Regional Memory Reader.}
The proposed Regional Memory Reader performs regional feature matching in both the memory and query frames, which reduces the number of mismatched pixels and therefore saves computational time.
To evaluate the effectiveness and efficiency of Regional Memory Reader, we replace the regional feature matching with global matching in the query frame and memory frame.
Compared to the memory readers that adopt global matching for memory or query frames, the proposed Regional Memory Reader achieves better results in terms of both accuracy and efficiency.
As shown in Figure~\ref{fig:mask-bbox-dist}, the areas of the regions containing target objects are usually smaller than 20\% of the whole image. Table~\ref{tab:ablation-regional-memory-reader} shows that the local-to-local matching in Regional Memory Reader is about 5 times faster (around 25 times smaller in FLOPS) than the global-to-global matching.
Figure~\ref{fig:ablation-regional-memory-reader} presents the visualization of the similarity scores of the target object in the memory readers, where the target object is highlighted in a red bounding box.
Given the estimated mask of the target object in the query frame, we compute the similarity scores in the previous frame, where the label of the target object in the query frame is determined by the pixels with high similarities.
Similarly, the pixels with similarity scores in the query frame are assigned with the label of the target object in the previous frame.
As shown in Figure~\ref{fig:ablation-regional-memory-reader}, the proposed Regional Memory Reader avoids the mismatching in the regions outside the target object in memory and query frames, which obtains better segmentation results.

\begin{table}
  \caption{The effectiveness of Regional Memory Reader. ``M.R.'' and ``Q.R.'' denote for ``Memory Region'' and ``Query Region'', where $\checkmark$ and $\times$ represent the feature matching is regional or global for the frame, respectively. The time for feature matching is measured on an NVIDIA Tesla V100 GPU without I/O time.}
  \vspace{.5 mm}
  \begin{tabularx}{\linewidth}{YY|ccYc}
    \toprule
    M.R.                 & Q.R.
    & $\mathcal{J}$ Mean & $\mathcal{F}$ Mean 
    & Avg.               & Time (ms) \\
    \midrule
    $\times$             & $\times$     & 
    0.792                & 0.843        & 0.818      & 10.68 \\
    $\checkmark$         & $\times$     &
    0.798                & 0.847        & 0.822      & 5.50 \\
    $\times$             & $\checkmark$ & 
    0.803                & 0.853        & 0.828      & 5.50 \\
    $\checkmark$         & $\checkmark$ & 
    \bf{0.810}           & \bf{0.860}   & \bf{0.835} & \bf{2.09} \\
    \bottomrule
  \end{tabularx}
  \label{tab:ablation-regional-memory-reader}
  \vspace{-2 mm}
\end{table}

\begin{table}
  \caption{The effectiveness of the ``Flow-based Region'' compared to ``Previous Region'' and ``Best-match Region'' used in FEELVOS~\cite{DBLP:conf/cvpr/VoigtlaenderCSA19} and KMN~\cite{DBLP:conf/eccv/SeongHK20}, respectively.}
  \vspace{.5 mm}
  \begin{tabularx}{\linewidth}{lYYY}
    \toprule
    Method            & $\mathcal{J}$ Mean & $\mathcal{F}$ Mean & Avg. \\
    \midrule
    Previous Region   & 0.762              & 0.822              & 0.792 \\
    Best-match Region & 0.792              & 0.845              & 0.819 \\
    \midrule
    Flow-based Region & \bf{0.810}         & \bf{0.860}         & \bf{0.835} \\
    \bottomrule
  \end{tabularx}
  \label{tab:ablation-region-predictor}
  \vspace{-2 mm}
\end{table}

\begin{table}
  \caption{The effectiveness of TinyFlowNet compared to FlowNet2-CSS~\cite{DBLP:conf/cvpr/IlgMSKDB17} and RAFT~\cite{DBLP:conf/eccv/TeedD20} for optical flow estimation. The time for optical flow estimation is measured on an NVIDIA Tesla V100 GPU without I/O time.}
  \vspace{.5 mm}
  \resizebox{\linewidth}{!}{
    \begin{tabularx}{1.05\linewidth}{lcccc}
      \toprule
      Method      & $\mathcal{J}$ Mean & $\mathcal{F}$ Mean 
                  & Avg.               & Time (ms) \\
      \midrule
      FlowNet2-CSS 
                  & \bf{0.814}         & \bf{0.860} 
                  & \bf{0.837}         & 59.93 \\
      RAFT 
                  & 0.808              & 0.859
                  & 0.834              & 157.78 \\
      \midrule
      TinyFlowNet & 0.810              & \bf{0.860}
                  & 0.835              & \bf{10.05} \\
      \bottomrule
    \end{tabularx}
  }
  \label{tab:ablation-tinyflownet}
  \vspace{-4 mm}
\end{table}

\noindent \textbf{Query Region Prediction.}
In RMNet, the regions for feature matching in the query frame are determined by the previous mask and the estimated optical flow.
In FEELVOS~\cite{DBLP:conf/cvpr/VoigtlaenderCSA19}, the regions for local feature matching are a local neighborhood of the locations where the target objects appear in the previous frame.
KMN~\cite{DBLP:conf/eccv/SeongHK20} performs feature matching in the regions determined by a 2D Gaussian kernel whose center is the best-matched pixel with the highest similarity score.
To evaluate the effectiveness of our ``Flow-based Region'', we compare its performance with different regions used for feature matching.
In Table~\ref{tab:ablation-region-predictor}, ``Previous Region'' and ``Best-match Region'' represent the regions determined by the methods used in FEELVOS and KMN, respectively.
As shown in Table~\ref{tab:ablation-region-predictor}, ``Flow-based Region'' outperforms ``Previous Region'' and ``Best-match Region'' in segmentation.
``Previous Region'' is based on the assumption that the motion
between two frames is usually small, which is not robust to object occlusion and drifting.
``Best-match Region'' only considers the region determined by the best-matched pixels.
However, the best-matched pixels are easily affected by lighting conditions and may be wrong for similar objects.

\noindent \textbf{TinyFlowNet.}
TinyFlowNet is designed to estimate  optical flow between two adjacent frames.
To evaluate the effectiveness of the proposed TinyFlowNet, we replace the TinyFlowNet with FlowNet2-CSS~\cite{DBLP:conf/cvpr/IlgMSKDB17} and RAFT~\cite{DBLP:conf/eccv/TeedD20} that are pretrained on the FlyingChairs~\cite{DBLP:conf/iccv/DosovitskiyFIHH15}.
As shown in Table~\ref{tab:ablation-tinyflownet}, the segmentation accuracies are almost the same when TinyFlowNet is replaced with FlowNet2-CSS and RAFT, which indicates that TinyFlowNet meets the need of region prediction in the query frame.
Moreover, the proposed RMNet with TinyFlowNet is 6 and 16 times faster than with FlowNet2-CSS and RAFT, respectively.

\section{Conclusion}

In this paper, we propose Regional Memory Network (RMNet) for semi-supervised VOS.
Compared to the STM-based methods, RMNet memorizes and tracks the regions containing target objects, which effectively alleviates the ambiguity of similar objects and also reduces the computational complexity for feature matching.
Experimental results on DAVIS and YouTube-VOS indicate that the proposed method outperforms the state-of-the-art methods with much faster running speed.

\noindent \textbf{Acknowledgements}
This work is conducted in collaboration with SenseTime. This work is in part supported by A*STAR through the Industry Alignment Fund - Industry Collaboration Projects Grant, in part by National Natural Science Foundation of China (Nos. 61772158, 61872112, and U1711265), and in part by National Key Research and Development Program of China (Nos. 2018YFC0806802 and 2018YFC0832105).

{\small
\bibliographystyle{ieee_fullname}
\bibliography{references}
}

\end{document}